\documentclass[letterpaper, 10 pt, journal, twoside]{ieeetran}
\usepackage{times}
\usepackage{graphicx}
\usepackage{subfigure}
\usepackage{amsmath}
\usepackage{amssymb}
\usepackage{amsfonts}
\usepackage{comment}
\usepackage{color}
\usepackage{multicol}
\usepackage{booktabs}
\usepackage{cite}
\usepackage{makecell}
\usepackage{multirow}  
\usepackage{notoccite}
\usepackage[bookmarks=true]{hyperref}
\usepackage[font=small,labelfont=bf]{caption}
\usepackage[top=1in, left=0.75in, bottom=0.75in, right=0.75in]{geometry}

\usepackage{pifont}

\IEEEoverridecommandlockouts
\begin{document}

\title{SphereVLAD++: Attention-based and Signal-enhanced Viewpoint Invariant Descriptor}

\markboth{IEEE Robotics and Automation Letters. Preprint Version. July, 2022}
{Zhao \MakeLowercase{\textit{et al.}}: SphereVLAD++: Attention-based and Signal-enhanced Viewpoint Invariant Descriptor}

\author{Shiqi Zhao$^\dagger$, Peng Yin$^{\dagger,*}$, Ge Yi, and Sebastian Scherer

\thanks{
This work was funded by US ARL award W911QX20D0008.

$^{1}$Shiqi Zhao is with University of California San Diego, La Jolla, CA 92093, USA. {(s2zhao@eng.ucsd.edu)}.

$^{2}$Peng Yin, Ge Yi, and Sebastian Scherer are with Robotics Institute, Carnegie Mellon University, Pittsburgh, PA 15213, USA {(pyin2, yige, basti@andrew.cmu.edu)}.

$^\dagger$ Shiqi Zhao and Peng Yin contribute equally in this work.

$^*$ Corresponding author: Peng Yin (hitmaxtom@gmail.com).

}
}

\maketitle
\begin{abstract}
LiDAR-based localization approach is a fundamental module for large-scale navigation tasks, such as last-mile delivery and autonomous driving, and localization robustness highly relies on viewpoints and 3D feature extraction.
Our previous work provides a viewpoint-invariant descriptor to deal with viewpoint differences; however, the global descriptor suffers from a low signal-noise ratio in unsupervised clustering, reducing the distinguishable feature extraction ability.
In this work, we develop SphereVLAD++, an attention-enhanced viewpoint invariant place recognition method.
SphereVLAD++ projects the point cloud on the spherical perspective for each unique area and captures the contextual connections between local features and their dependencies with global 3D geometry distribution.
In return, clustered elements within the global descriptor are conditioned on local and global geometries and support the original viewpoint-invariant property of SphereVLAD.
In the experiments, we evaluated the localization performance of SphereVLAD++ on both the public \textit{KITTI360} dataset and self-generated datasets from the city of Pittsburgh.
The experiment results show that SphereVLAD++ outperforms all relative state-of-the-art 3D place recognition methods under small or even totally reversed viewpoint differences and shows 7.06\% and 28.15\% successful retrieval rates with better than the second best.
Low computation requirements and high time efficiency also help its application for low-cost robots.
\end{abstract}


\begingroup
\let\clearpage\relax
    \section{Introduction}

    \IEEEPARstart{L}{ocalization} is one of the essential parts in long-term navigation of large-scale environments, such as campus and urban areas.
    Based on LiDAR’s natural advantage over visual input in terms of illumination influence, successful point-based localization approaches have been widely applied in autonomous driving~\cite{Intro:automous1} and last mile delivery.
    However, except the sensitivity of LiDAR-based methods to the viewpoint differences, the Signal-Noise Ratio (SNR) of global descriptors in challenging 3D environments is still at low level.
    Although global descriptors are represented in a high-dimensional feature space, the number of channels that contribute to the task is usually limited.
    Consequently, the place recognition (or loop-closure detection) ability will be reduced along with the complexity of the 3D environments.
    
    \begin{figure}[t]
    	\centering
        \includegraphics[width=\linewidth]{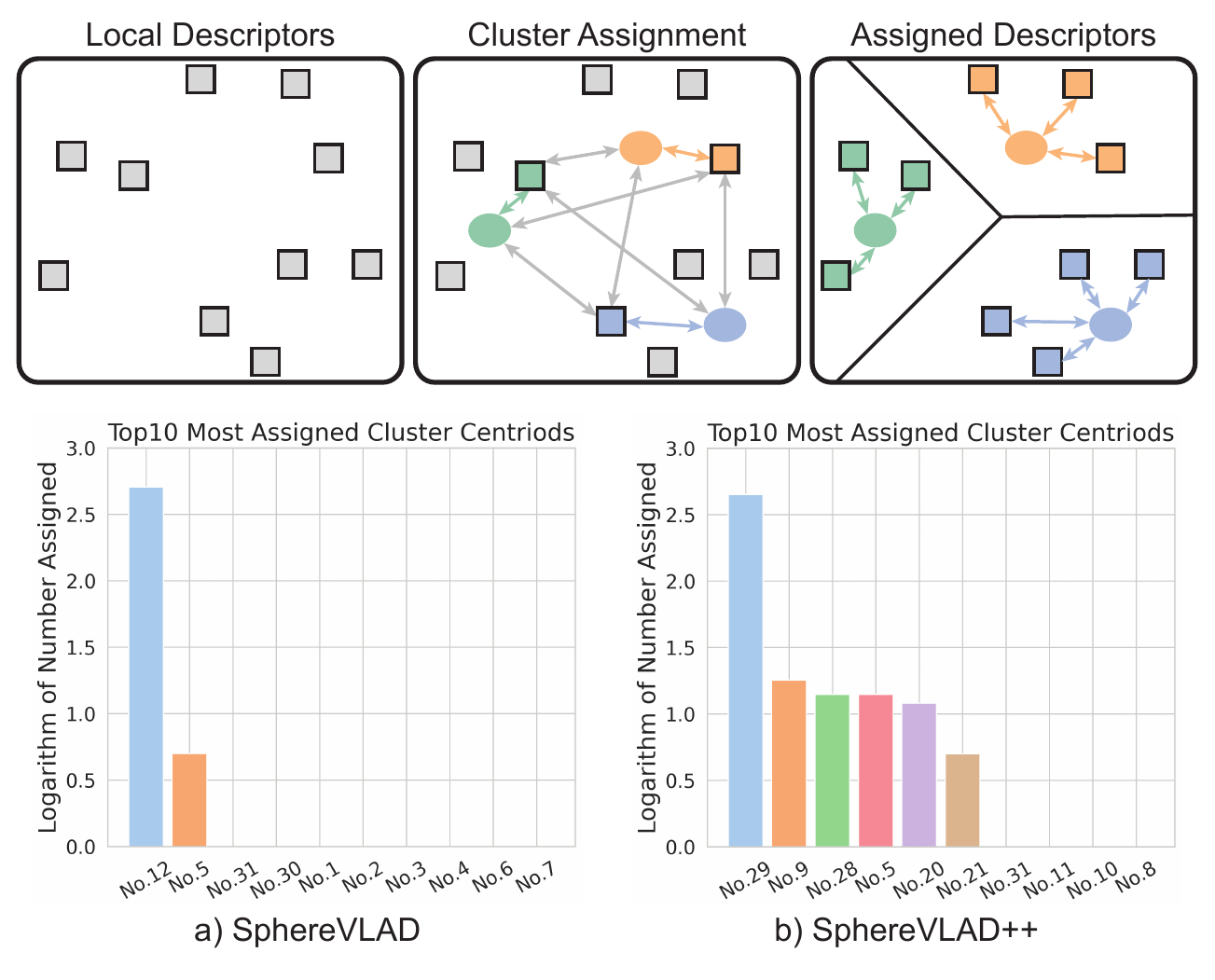}
    	\caption{\textbf{SphereVLAD++} is novel viewpoint invariant LiDAR place recognition method which exploits the relationships between local features. By enhancing the local features with contextual information, we are able to cluster more active centroids and improve the soft-assignment performance in NetVLAD layer. Compared with SphereVLAD, SphereVLAD++ has more active centroids and better local descriptor assignment distribution.}
    	\label{fig:idea}
    \end{figure}
    
    In the current point-based place recognition approaches, many existing works~\cite{LPR:PointNetVLAD, PR:LPDNet} encode the global place descriptor via a point-wise~\cite{PointNet, pointnet++} or voxel-wise~\cite{voxnet} feature extractor.
    The major disadvantage of point- and voxel- wise approaches roots in their sensitivity to viewpoint differences.
    As demonstrated in LPDNet~\cite{PR:LPDNet}, the PointNet++~\cite{pointnet++} enhanced place descriptor can further improve the robustness to small viewpoint differences (orientation within $30^\circ$).
    However, the viewpoint differences of real world applications is way more complicate, especially for open areas, and the performance of the above methods drops dramatically when facing significant viewpoint differences.
    More recently, Xia~\emph{et al.} proposed SOE-Net~\cite{PR:SOENet}, which employs an orientation encoding unit for PointNet to learn local neighborhood in different orientations.
    Nevertheless, extracting point-based feature encoding in limited number of directions is computation expensive and cannot fully exploit a complete rotation invariant features.
    Instead of point-based, Chen~\emph{et al.} introduced a projection based recognition~\cite{PR:OverlapNet}~\cite{PR:OverlapNet_jour}, which estimates localization and relative orientation based on the overlaps of testing and reference frames.
    However, the orientation-invariant property is highly relying on the training data configurations, and shows limited generalization ability for unseen environments.
    
    Attention mechanism has been proved as a great success in natural language process (NLP) and computer version, and few works~\cite{PR:SOENet,LPR:PCAN,LPR:DAGC,barros2021attdlnet} have introduced attention mechanism into their networks for LiDAR-based place recognition.
    Alike the attention of human, attention mechanism learns the patterns between local features and therefore balances the importance among them, which improves features' distinguishability for recognition.
    PCAN~\cite{LPR:PCAN} firstly proposed a contextual attention network which extracts the patterns of the points within a neighborhood to reconsider the significance of each local feature during global descriptor aggregation\cite{PR:netvlad}.
    However, re-weighting the importance among local features cannot handle the viewpoint differences, and the contextual information are limited by the quantity of the points within the neighborhood.
    
    In our previous work~\cite{LPR:seqspherevlad}, SphereVLAD, we introduced an viewpoint-invariant descriptor.
    Instead of extracting features from point cloud directly, the point cloud is converted into panorama via spherical projection, and viewpoint-invariant descriptors are obtained via investigating the spherical harmonics.
    In this work, as an extension of~\cite{LPR:seqspherevlad}, we propose SphereVLAD++, an attention-based and signal-enhanced viewpoint invariant 3D descriptor.
    The proposed attention module can learn the inner relationships among local features and local-to-global connections.
    Finally, the place descriptors are clustered conditioned on the new local features which captures both local and global connections simultaneously.
    Our main contributions can be summarized as following:
    \begin{itemize}
        \item We develop an attention based viewpoint-invariant place descriptor, SphereVLAD++, which can enhance the local features with long-range dependencies through attention mechanism. The extracted global descriptor is conditioned on both local and global contexts, which can further improve the distinguishability under variant viewpoints.
        \item We provide a Signal-Noise Ratio metric to evaluate the performance of the NetVLAD layer based on the number of cluster centroids contributing to the global descriptor. 
        We conduct comparison experiments for SphereVLAD and SphereVLAD++ in both seen and unseen environments.
        \item We conduct the place recognition evaluations for point-based approaches, and provide detailed accuracy and efficiency analysis of different approaches, and investigate the generalization ability of SphereVLAD++ with and without attention module.
    \end{itemize}
    
    In the experiment results, we present an extensive evaluation of our system on three unique datasets: 
    (1) the \textit{Pittsburgh} dataset consists of $50$ trajectories collected in the city of Pittsburgh; 
    (2) the \textit{KITTI360} dataset consists of $11$ individual trajectories recored in suburbs of Karlsruhe;
    (3) and the \textit{Campus} dataset consists of $10$ trajectories with overlaps within the campus of Carnegie Mellon University (CMU).
    Leveraging the attention enhancement module helps our previous SphereVLAD method achieve strong stability in large translation differences and attain higher generalization ability for unseen environments.
    We include a discussion section and a conclusion section to analyze the advantages/disadvantages and future works for our SphereVLAD++ descriptor.

\section{Related Work}
\label{sec:related_work}

        Base on the representations of point cloud, LiDAR-based place recognition can be summarized into three categories: point-, voxel- and projection- based place recognition methods.
        For point-based approach, Mikaela~\textit{et al}~\cite{LPR:PointNetVLAD} utilize PointNet~\cite{PointNet} to extract local features, and cluster them into a global place descriptor via the deep vector of locally aggregated descriptors (VLAD)~\cite{PR:netvlad}.
        This work enables the direct 3D place feature learning from point-cloud.
        But PointNetVLAD omits relationships between local features, which will significant reduce the localization accuracy when under different viewpoints.
        Based on extended 3D point feature extraction of PointNet++~\cite{pointnet++}, LPD-Net~\cite{PR:LPDNet} takes both point cloud and handcrafted features as input and introduces a graph-based aggregation module following the PointNet++ framework to learn multi-scale spatial information. 
        This method improves orientation sensitivity of PointNetVLAD, but still is fragile to large orientation differences.
        Moreover, SOE-Net~\cite{PR:SOENet} recently modifies PointNet by adding orientational feature embeddings, which learn local geometry from eight spatial orientations to reduce the sensitivity to orientation variances.
        However, SOE-Net can only cover limited spatial orientation differences, which also restrict its generalization ability for unseen environments.
        And all point-based feature extraction ability are highly restricted to small points numbers (i.e. $4096$), which may affect the localization accuracy under different resolutions and perspectives.
        For voxel-based approach, Sriram~\emph{et al.}~\cite{LPR:3DVox} provide a voxel-based 3D place recognition method, which shows the advantages in long-term environment variations.
        MinkLoc3D~\cite{LPR:minkloc3d} also uses sparse 3D convolutions on sparse voxelized point cloud to extract local features in a computationally effective way.
        Instead of extracting the features directly from the point cloud, RINet [17] proposed by Li~\emph{et al} utilizes high-level semantic information to enhance place recognition performance.
        For projection based approaches, Kim~\emph{et al.}~\cite{PR:ScanContext} propose a projection-based descriptor called Scan-Context to solve long-term global localization. 
        Instead of extracting the features directly from the point cloud, RINet~\cite{LPR:RINet} proposed by Li~\emph{et al.} utilizes high-level semantic information to enhance place recognition performance.
        Most recently, Chen~\emph{et al.}~\cite{PR:OverlapNet}~\cite{PR:OverlapNet_jour} introduce an overlapping estimation network to predict the place feature difference and the relative yaw differences simultaneously.
        
        In our previous work SphereVLAD~\cite{LPR:seqspherevlad}, we provided an viewpoint-invariant place descriptor based on the spherical harmonics, which is not only robust to viewpoint differences but a light-weight network model.
        However, SphereVLAD doesn't pay much attention to the long-range dependencies between local features and one promising solution is to use the attention mechanism. 
        Attention mechanism has been widely used in computer vision~\cite{attention:swin,attention:vit} related tasks including classification and segmentation.
        The long-range contextual information generated from attention mechanism provides local descriptors with correlation information.
        Recently, researchers have introduced attention mechanism into 3D place recognition.
        In PCAN~\cite{LPR:PCAN}, the author produces a attention map to predict the significance of each local point feature based on point context.
        SOE-Net~\cite{PR:SOENet} uses a self-attention unit to encode meaningful spatial information between local features.
        And DAGC~\cite{LPR:DAGC} employs a dual-attention module to extract local features from point-cloud directly with one module focusing on weights between points and the other module focusing on weights within each feature channel.
        Recently, PPT-Net~\cite{PR:PPTNet} and OverlapTransformer~\cite{LPR:overlaptransformer} use the transformer network to improve the descriptiveness of descriptors.
        Inspired by \cite{zhang2019self}, SphereVLAD++ extends our previous work~\cite{LPR:seqspherevlad} with an light-weight attention module to enhance the representation ability of local features. 
        In the experiment results, we can show the significant improvement in dealing with large translation differences from SphereVLAD to SphereVLAD++ and better generalization ability for unknown environments.
        
    \section{Our Method}
\label{sec:system_overall}
    \begin{figure*}[ht]
        \centering
        \includegraphics[width=\linewidth]{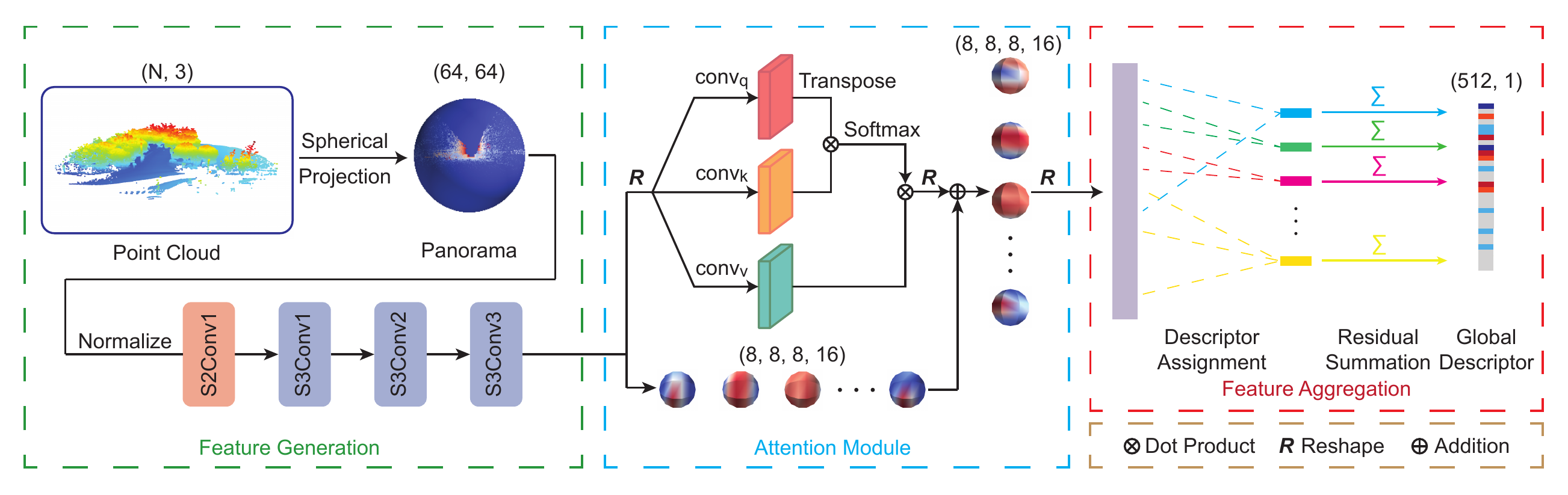}
        \caption{\textbf{Network Structure of SphereVLAD++.}
        The network takes a query LiDAR scan as input and then convert it into panorama by spherical projection. Then local oorientation-equivalent features are encoded by spherical convolution and added with contextual information generated from attention module. Finally, NetVLAD layer is utilized for descriptor aggregation and produces a global descriptor.
        }
        \label{fig:network}
    \end{figure*}
    In this section we will introduce the details of our framework.
    As shown in Fig.~\ref{fig:network}, the point cloud is projected onto the spherical perspectives for orientation-equivalent local feature extraction.
    Then SphereVLAD++ feeds extracted local features to the attention module, which will exploit the correlation between local features and yield a weighted combination of original local features and contextual information.
    Finally, the extracted global descriptor is clustered conditioned on the new attention enhanced features with the VLAD layer.
    In the following subsections, we will investigate the three modules and loss metrics respectively.

    \subsection{Spherical Feature Extractor}
    \label{sec:SphereVLAD}
    To learn viewpoint-invariant place features, our method utilizes the spherical convolutions for orientation-invariant feature extraction.
    As shown in Fig.~\ref{fig:encoder}, our feature extractor contains 4 convolution layers.
    The first layer performs a convolution on $S^2$ and the remaining three layers perform on $\mathbf{SO}(3)$.
    Furthermore, we apply a ReLu activation function and a three-dimensional batch normalization after each spherical convolution.
    SphereVLAD++ first project the 3D point cloud into spherical coordinates $I \in S^2$, which is constraints with $\alpha \in [0,2\pi]$ in horizontal and $\beta \in [0, \pi]$ in vertical.
    $I$ is modeled as continuous function $f: S^2 \rightarrow \mathbb{R}^K$, which is denoted as spherical signal.
    Given two spherical signals $f$ and $\phi$, spherical correlation is defined as:
    \begin{align}
        [\phi \star f](\mathbf{R}) = & \int_{S^2}\phi(\mathbf{R}^{-1}I)\psi(I)d\mathbf{Q}.
    \end{align}
    Furthermore, spherical correlation is proved to be held invariant between rotations. 
    As the output of spherical correlation is a function on $\mathbf{SO}(3)$ which is denoted as $\mathbf{SO}(3)$ signal, given $g, \psi \in \mathbf{SO}(3) \rightarrow \mathbb{R}^K$,
    correlation between $g$ and $\psi$ is defined as:
    \begin{align}
        [\psi \star_{\mathbf{SO}(3)} g](\mathbf{R}) = & \int_{\mathbf{SO}(3)}\psi(\mathbf{R}^{-1}\mathbf{Q})g(\mathbf{Q})d\mathbf{Q}.
    \end{align}
    where $\mathbf{R, Q} \in \mathbf{SO}(3)$. 
    Based on the proof in~\cite{cohen2018spherical}, spherical correlation is shown to be orientation-equivalent:
    \begin{align}
        [\psi \star_{\mathbf{SO}(3)} [L_{\mathbf{G}}\psi]](\mathbf{R})
        = & [L_{\mathbf{G}}[g \star_{\mathbf{SO}(3)} \psi]](\mathbf{R})           
    \end{align}
    where $L_{\mathbf{G}} (\mathbf{G} \in \mathbf{SO}(3))$ is the rotation operator for spherical signals.
    In practice, the computation of spherical correlation between two spherical signals $f$ and $\phi$ is shown as following. 
    First, spherical signals are expanded in spherical harmonics domain and point-wise product of harmonic coefficients are computed. 
    Finally, output is calculated by inverting the spherical harmonics expansion.
    Due to the expensive operation of spherical convolution~\cite{shapenet}, it's hard to design deep convolution layers to capture the connections between local and global geometric features.
    Our attention enhancement module can help deal with this problem.
    
    \begin{figure}[ht]
        \centering
        \includegraphics[width=\linewidth]{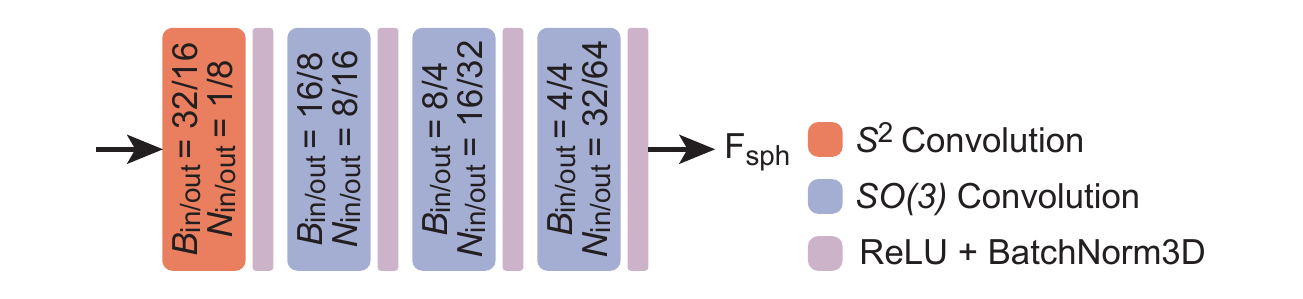}
        \caption{Detailed Structure of Spherical Feature Extractor
        }
        \label{fig:encoder}
    \end{figure}
    
    \subsection{Attention Enhanced Global Descriptor Generator}
    The expensive operation of spherical convolution also leads to small image input which can only provide limited details.
    Furthermore, local feature map extractor only focuses on the information in a limited local neighborhood, which leads to insufficiency when global descriptor generator learns the patterns from the local feature map.
    Therefore, we utilize an attention module which allows local descriptors to gain contextual information and further contributes to better performance in global descriptors extraction.

    \subsubsection{Self-Attention on Spherical Signals}
    Different from sentences or images, output $F_{sph}$ generated from spherical convolution is in the shape of $(C \times \alpha \times \beta \times \gamma)$, where $\alpha$, $\beta$, $\gamma$ are Euler angles parameterize rotation group $SO(3)$.
    As the input required by attention module is a sequence, given $F_{sph}$, we reshape it into $F \in \mathbb{R}^{C \times L}$, where ${L=\alpha \times \beta \times \gamma}$.
    Then $F$ is transformed into three different feature spaces\cite{zhang2019self}: query feature space $Q = W_q F$, key feature space $K = W_k F$ and value feature space $V = W_v F$, where $Q, K, V \in \mathbb{R}^{C^{\prime} \times L}$ and $W_q, W_k, W_v \in \mathbb{R}^{C^{\prime} \times C}$. 
    Contextual information $A \in \mathbb{R}^{L \times L}$ therefore is generated as following:
    \begin{align}
        \label{eq:attention}
        A = \text{softmax}(Q K^{T})V
    \end{align}
    Once contextual information $A$ is obtained, it will be multiplied with a trainable parameter $\omega$ and then added with original local feature map $F$ to yield a new feature map $F^{\prime}$.
    \begin{align}
    \label{eq:attention_sum}
        F^{\prime} = F + \omega A
    \end{align}
    Since the contextual information is utilized to modify the original feature map, $\omega$ is trained to balance the importance between the feature map and contextual information.
    Based on Eq.~\ref{eq:attention_sum} and Eq.~\ref{eq:attention}, each local descriptor $f^{\prime}_i$ can be interpreted as a summation of $f_i$ and a weighted sum of other local descriptors.
    The weighted sum can be further regarded as semantic information and $W^A_j$ reveals the contribution of each local descriptor $f_j$ to $f_i$.
    \begin{align}
    \label{eq:attenion_semantic}
        f^{\prime}_i = f_i + \sum^{N}_{j} W^A_j f_j
    \end{align}
    Moreover, the summation enlarges the receptive field, which alleviates the shortage of shallow spherical convolution encoder.
    Our attention module exploits the spatial relations between local descriptors and provides semantic information, which enhances the performance of feature extraction.
    
    \subsubsection{Attention Enhanced NetVLAD}
    \label{method:attention_vlad}
    The global descriptor generator is chosen as VLAD layer \cite{PR:netvlad}, which clusters local descriptors into $K$ cluster centroids and utilizes a soft-assignment mechanism to assign local descriptors to corresponding cluster centroid(Eq.~\ref{eq:vlad}). Intuitively, the performance of local descriptor clustering and soft-assignment will be improved with the help of contextual information.
    \begin{align}
    \label{eq:vlad}
        V_k &= \sum_{i}^{N}a_k^i(x_i - c_k) \\
        \overline{a}_k^i &= \frac{\exp({w_k^T x_i + b_k})}{\sum_{k^{\prime}}\exp({w_{k^{\prime}}^T x_i + b_{k^{\prime}}})}
    \end{align}
    Given the fact of Eq.~\ref{eq:attenion_semantic}, soft-assignment for attention enhanced local descriptors can be rewritten as:
    \begin{align}
    \label{eq:soft_assign}
        \overline{a}_k^i = \frac{\exp({w_k^T (x_i+\sum^{N}_{j} W^A_j x_j) + b_k})}{\sum_{k^{\prime}}\exp({w_{k^{\prime}}^T (x_i+\sum^{N}_{j} W^A_j x_j) + b_{k^{\prime}}})}
    \end{align}
    Compared with the original soft-assignment function $S_{assign}(\overline{a}_k^i|x_i, c_1...c_K)$ which only considers the relations between $x_i$ and cluster centroids $c_1...c_K$, the attention enhanced function $S^{\prime}_{assign}(\overline{a}_k^i|x_1...x_N, c_1...c_K)$ takes all local descriptors and cluster centroids into account and balances the cluster assignment across each descriptor.
    
    \subsubsection{Viewpoint Invariance}
    Due to the property of spherical convolution mentioned in~\ref{sec:SphereVLAD}, our feature encoder can generate viewpoint invariant local features.
    In the attention module, all components including linear transformation, softmax function and addition along channels are also viewpoint invariant.
    For the global descriptor generation, \cite{LPR:PointNetVLAD} proves that NetVLAD is permutation invariant and \cite{LPR:overlaptransformer} proves that global descriptor which is permutation invariant is also viewpoint invariant if local features are viewpoint invariant.
    Therefore, our proposed method can produce viewpoint invariant global descriptors.
    
    \subsection{Implementation Details}
    For data preprocessing, we firstly generate submaps by accumulating LiDAR scans\cite{merge:automerge} and take the points within $50$ meters distance from the origin of each submap and project them into a panorama whose size is $64 \times 64$ using the method mentioned in \cite{LPR:fusionvlad}.
    To enable the end-to-end training for our SphereVLAD++ module, we utilize the "Lazy quadruplet" loss metric.
    Sets of training tuples $\mathcal{S} = [S_a, \{S_{pos}\}, \{S_{neg}\}, S_{neg}^{*}]$ are selected from the training dataset.
    Two frames are regarded as positive if their geometric distance is less than the threshold $D_{pos}$ and regarded negative if strictly larger than the threshold $D_{neg}$.
    In our case, these two thresholds are set as $D_{pos}=8m$ and $D_{neg}=16m$.
    The lazy quadruplet loss is defined as
    \begin{align}
        L_{lazyQuad}&(\mathcal{S}) = \\ \nonumber
        &\max_{i, j}([m_1 + d(g_a, g_{pos_i}) - d(g_a, g_{neg_j})]) + \\ \nonumber
        &\max_{i, k}([m_2 + d(g_a, g_{pos_i}) - d(g_{neg_k}, g_{neg^{*}})]) \nonumber
    \end{align}
    where $g$ is the corresponding global descriptors encoded by SphereVLAD++, and $d(\cdot)$ denoted the Euclidean distance. $m_1$ and $m_2$ are the constant threshold giving the margin. In our case, $m_1$ and $m_2$ are set to $0.5$ and $0.2$.


\section{Experiment.}
    In this section, we will investigate the localization accuracy of SphereVLAD++ against other state-of-the-art 3D place recognition methods over different outdoor datasets.
    We also evaluate the SNR rate of SphereVLAD and SphereVLAD++ for unseen environments, which can analysis the generalization ability of global descriptors in real applications and the improvement introduced by attention module.
    All the experiments are conducted on an AMD 5800x processor and a single Nvidia RTX2060 with 12G memory.
    In the rest of this section, we will introduce our evaluation datasets, the place recognition metrics and the SNR information rate, and the relative state-of-the-art methods.
    Then place recognition performance under viewpoint-differences will be evaluated, followed by the localization analysis on unseen environments.
    
\subsection{Dataset Overview}
    Our datasets include the public \textit{KITTI360} dataset and our self-collected datasets:
    \begin{itemize}
        \item \textbf{Pittsburgh}~\cite{dataset:alita}. 
        We created a City-scale dataset by mounting our data capturing system on the top of a car and incrementally traversing $200km$ in the city of Pittsburgh.
        The data-collection platform contains a LiDAR device (Velodyne-VLP $16$), an inertial measurement unit (Xsense MTi $30$, $0.5^{\circ}$ error in roll/pitch, $1^{\circ}$ error in yaw) and a mini Intel computer.
        This dataset is divided into $50$ trajectories, we use trajectory $\{1 \sim 15\}$ for training and trajectory $\{21 \sim 40\}$ for evaluation on the viewpoint-invariant and unseen environment analysis.
        
        \item \textbf{Campus}~\cite{dataset:alita}.
        We recorded within the campus area of Carnegie Mellon University (CMU) using the same data collection device with \textit{Pittsburgh}.
        This dataset covers $10$ main trajectories throughout the campus and the total length is around $36km$.
        Each trajectory is recorded 4 times in both forward and backward directions under different illumination conditions.
        $\{1 \sim 6\}$ tracks are used for viewpoint-invariant analysis.
        
        \item \textbf{KITTI360}~\cite{DATASET:KITTI360} consists of $11$ individual sequences generated with a Velodyne HDL-64E LiDAR device covering $73.7km$ of the suburban areas of Karlsruhe, Germany.
        The dataset has 6 sequences (0000, 0002, 0004, 0005, 0006, 0009) with loop closures, which are used for unseen environment analysis.
    \end{itemize}
    
    
    In \textit{Pittsburgh} dataset, query frames and database frames are generated within each trajectory. 
    We collect the key poses from traditional LiDAR odometry~\cite{LOAM:zhang2014loam} and choose the key poses around every $5$ meters. 
    The corresponding submaps of these key poses are treated as database frames and submaps corresponding to other key poses are regarded as query frames.
    In \textit{KITTI360} dataset, we find all the revisits within the same trajectory and treat the submaps of revisiting key poses as query frames and all others as database frames.
    In \textit{Campus} dataset, as each trajectory is recorded 8 times, we utilize interactive SLAM~\cite{interactiveslam} to find the geometric relations between the poses of different recordings.
    We collect the key poses every $3$ meters and the first recording in forward direction is used to generate database frames.
    The second recording in the forward direction and the first recording in the reserve direction are utilized to generate query frames.

\subsection{Evaluation Metrics and Comparison Methods}
\label{sec:result}

    \begin{table*}[t]
        \centering
        \caption{Average Recall ($\%$) on Three Datasets.}
        \label{TB:recall}
        \begin{tabular}{c c c c c c c c c}
        \toprule
                          &\textit{Pittsburgh(@1)} &\textit{KITTI360(@1)} &\textit{Campus(@1)} &\textit{Pittsburgh(@1\%)} &\textit{KITTI360(@1\%)} &\textit{Campus(@1\%)}\\ 
        \midrule
        PointNetVLAD~\cite{LPR:PointNetVLAD}         & $69.66$  & $24.05$  & $49.09$  & $70.50$  & $32.26$  & $57.75$\\
        PCAN~\cite{LPR:PCAN}                         & $70.55$  & $26.10$  & $54.65$  & $71.97$  & $30.79$  & $64.81$\\
        LPD-Net~\cite{PR:LPDNet}                     & $73.06$  & $25.51$  & $54.33$  & $74.16$  & $31.06$  & $65.98$\\
        SOE-Net~\cite{PR:SOENet}                     & $77.14$  & $63.10$  & $45.34$  & $78.56$  & $\mathbf{95.48}$  & $55.05$\\
        MinkLoc3D~\cite{LPR:minkloc3d}               & $84.05$  & $45.45$  & $\mathbf{56.26}$  & $84.94$  & $66.63$  & $\mathbf{75.83}$\\
        SphereVLAD~\cite{LPR:seqspherevlad}          & $61.76$  & $40.47$  & $33.48$  & $62.76$  & $59.53$  & $39.78$\\
        SphereVLAD++(ours)       & $\mathbf{91.11}$  & $\mathbf{68.04}$  & $67.48$ & $\mathbf{91.58}$  & $83.87$  & $75.08$\\
        \bottomrule
        \end{tabular}
    \end{table*}

    We compare with state-of-the-art learning based 3D place recognition methods to evaluate the performance of our method. 
    PointNetVLAD~\cite{LPR:PointNetVLAD} LPD-Net~\cite{PR:LPDNet}, PCAN~\cite{LPR:PCAN}, SOE-Net~\cite{PR:SOENet}, MinkLoc3D~\cite{LPR:minkloc3d} and SphereVLAD~\cite{LPR:seqspherevlad} are chosen by taking advantages of their official implementations on Github.
    For point-based or voxel-based methods, non-informative ground planes of each submap are removed and the resulting point cloud is downsampled to 4096 points.
    We utilize Recall@$N$ to show the performance of each method, and Average Recall@$1$(AR@$1$) and Average Recall@$1\%$(AR@$1\%$) are selected to compare the retrieval and generalization ability of each method.
    Successful retrieval is define as retrieving a point cloud within $5m$ for \textit{Pittsburgh} and \textit{KITTI360} dataset and $3m$ for \textit{Campus} dataset.
    
\subsection{Localization Accuracy Analysis}
\label{sec:accuracy}
    For \textit{Pittsburgh} dataset, we generate query frames with random translation noise sampled in range of $[-3m \sim 3m]$ and random yaw orientation noise sampled in range of $[-30^{\circ} \sim 30^{\circ}]$.
    As shown in Table.~\ref{TB:recall}, our attention module shows great performance improvement on Pittsburgh dataset.
    SphereVLAD++ outperforms SphereVLAD by 29.35\%, which greatly alleviates the inferiority in dealing with large translation differences.
    Our method also surpasses point-based methods by a margin, which proves the stability under small rotation noise.
    For unseen datasets, the results show the superior generalization ability of our method.
    As \textit{KITTI360} contains both forward and reverse loop closures, point-based methods including PointNetVLAD, PCAN and LPD-Net are greatly affected by the viewpoint changing.
    The $27.57\%$ enhancement between SphereVLAD and SphereVLAD++ shows the contextual information extracted from our attention module is not outfitted to \textit{Pittsburgh} and can efficiently capture task-relevant long-term dependencies between local features in the unseen environments.
    We further examine the performance of relocalization using \textit{Campus} dataset.
    Both database and query frames are selected in the trajectories in same direction and the distance between adjacent database frames is around $3m$.
    Compared with other methods, SphereVLAD++ outperforms them on AR@$1$ but inferior to the state-of-art method MinkLoc3D on AR@$1\%$.
    The ability to generate more distinguishable features to retrieve the correct database frames in densely located situations can realize more precise robot self-localization.

\subsection{Viewpoint-invariant Analysis}
\label{sec:viewpoint}
    To further figure out the performance of place recognition under different viewpoints, we conduct experiments on \textit{Pittsburgh} with different yaw orientations $[0^{\circ},30^{\circ},60^{\circ},90^{\circ},120^{\circ},150^{\circ},180^{\circ}]$ and random translation noise sampled in range of $[-1m \sim 1m]$.
    As shown in Fig.~\ref{fig:viewpoint}, PointNetVLAD, PCAN and LPD-Net drop quickly as the orientation difference increases.
    SOE-Net and MinkLoc3D is not significantly affected under certain viewpoint differences but cannot achieve viewpoint invariance.
    On the contrary, SphereVLAD and SphereVLAD++ remain consistence in increasing viewpoint changing.
    Notably, the robustness in translation noise of SphereVLAD++ is enhanced by a obvious margin compared with SphereVLAD.
    \begin{table}[ht]
        \label{exp:table_viewpoint_campus}
        \centering
        \caption{Average Recall on Campus Dataset(backward)}
        \begin{tabular}{c c c c}
        \toprule
        Method       &  Recall@1            & Recall @1\% \\
        \midrule
        PointNetVLAD &     $3.17$           & $4.49$  \\
        LPD-Net      &     $3.95$          & $4.98$ \\
        PCAN         &     $3.75$           & $4.83$ \\
        SOE-Net      &     $28.37$          & $39.65$ \\
        Minkloc3D    &     $26.61$          & $36.91$ \\
        SphereVLAD   &     $26.18$          & $31.87$ \\
        SphereVLAD++(ours) & $\mathbf{56.52}$  & $\mathbf{64.95}$ \\
        \bottomrule 
        \label{table:time_analysis}
        \end{tabular}
    \end{table}
    
    In addition, we also test the performance of reverse loop detection in \textit{Campus}.
    Reverse loop detection is really common in real life scene and therefore fundamental in large-scale autonomous driving.
    Database frames remain the same with experiments in Sec.~\ref{sec:accuracy} but query frames are selected in the trajectories in the reverse direction.
    Compared with results shown in Table.~\ref{TB:recall}, we can see obvious decrease for point-based and voxel-based methods.
    SphereVLAD and SphereVLAD++ maintain the performance with only small decline of $7.30\%$ and $10.96\%$ respectively.
    \begin{figure}[t]
        \centering
        \includegraphics[width=\linewidth]{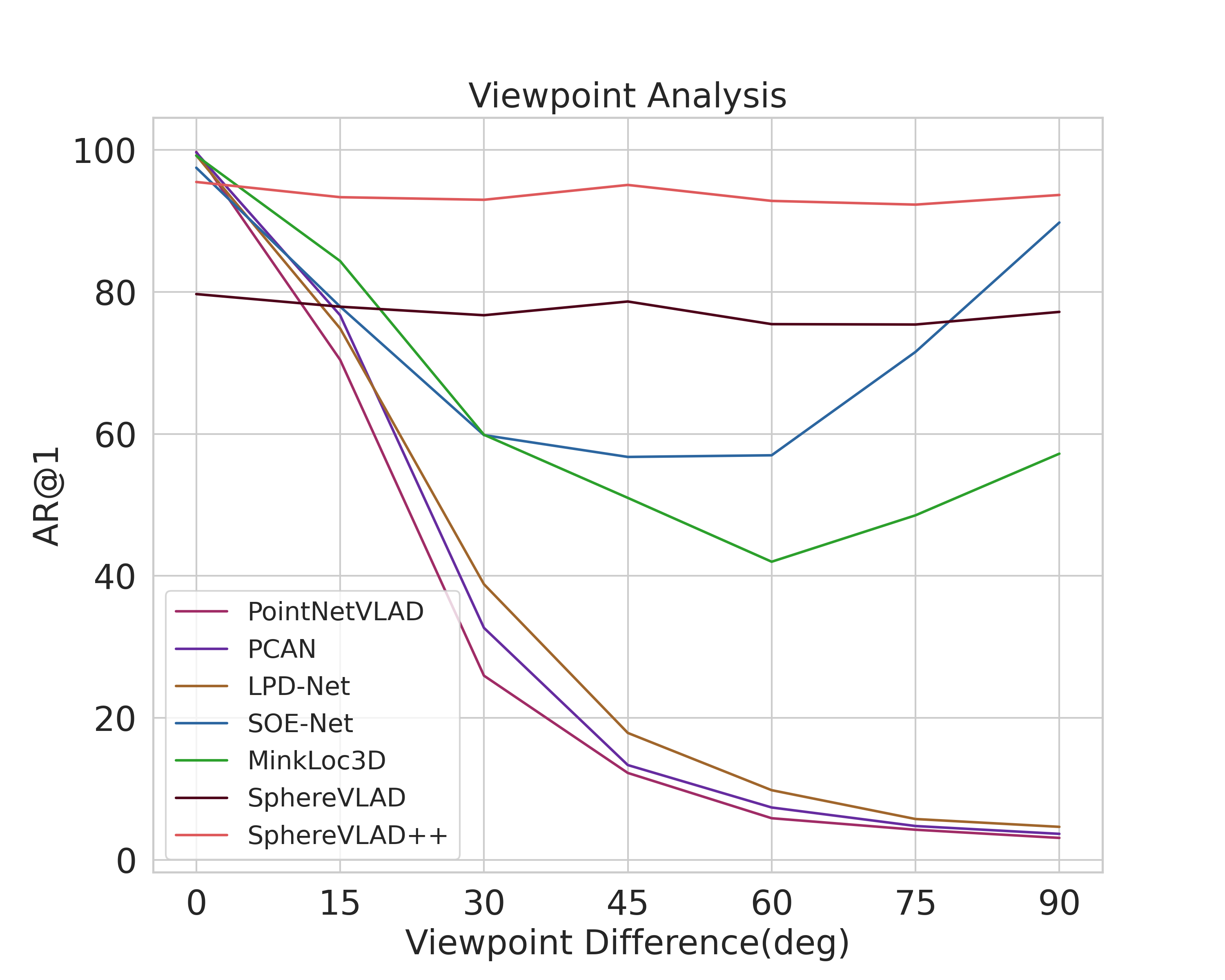}
        \caption{AR@$1$ of Methods under Different Viewpoints.}
        \label{fig:viewpoint}
    \end{figure}

\subsection{Signal Ratio Noise Analysis}
    As mentioned in sec.~\ref{method:attention_vlad}, the performance of the soft-assignment is influenced by the contextual information generated from attention module.
    Therefore, we conduct experiments to inspect the variables inside the NetVLAD layer in both training and unseen datasets.
    As NetVLAD layer uses a soft-assignment mechanism demonstrated in Eq.~\ref{eq:soft_assign} to classify each local descriptor to a specific cluster centroid, we save the soft-assignment variable $\overline{a}$ to evaluate the performance.
    We find out that even though cluster centroid number set to NetVLAD layer during training is 32, the network does not exactly cluster the local descriptors as expected.
    Local descriptors only response to limited number of cluster centroids which we define as active centroids.
    Inspired by original Signal Noise Ratio(SNR) which is defined as the ratio of the power of a signal (meaningful input) to the power of background noise (meaningless or unwanted input), Signal Noise Ratio(SNR) for NetVLAD layer is defined as:
    \begin{align}
    \label{eq:srn}
        \text{SNR} =\frac{N_{active}}{N_{total} - N_{active}}
    \end{align}
    where $N_{active}$ and $N_{total}$ denote the number of active centroids and total centroids respectively.
    As shown in Table.~\ref{exp:snr_anlysis}, the SNR rate of SphereVLAD++ is higher than SphereVLAD in all three datasets, which shows that contextual information can help to yield global descriptor with more details.
    
    \begin{table}[ht]
        \centering
        \caption{SNR Rate on Different Datasets}
        \begin{tabular}{c c c c c c c}
        \toprule
        Method       &  Pittsburgh  & KITTI360 & Campus \\
        \midrule
        SphereVLAD   & $0.143$  & $0.103$ & $0.103$ \\
        SphereVLAD++ & $\mathbf{0.28}$  & $\mathbf{0.28}$ & $\mathbf{0.333}$ \\
        \bottomrule 
        \end{tabular}
        \label{exp:snr_anlysis}
    \end{table}

    Fig.~\ref{fig:snr_dataset} shows the cluster assignment of local descriptors generated from all samples in test set.
    It is obvious that both SphereVLAD and SphereVLAD++ have a dominating centroid which most local descriptors are assigned to.
    Moreover, Fig.~\ref{fig:snr_dataset} also shows the extra active centroids in SphereVLAD++ do contribute in cluster assignment.
    Based on different dataset, the active centroids trained in \textit{Pittsburgh} are also applicable in unseen environment.
    
    \begin{figure}[ht]
        \centering
        \includegraphics[width=\linewidth]{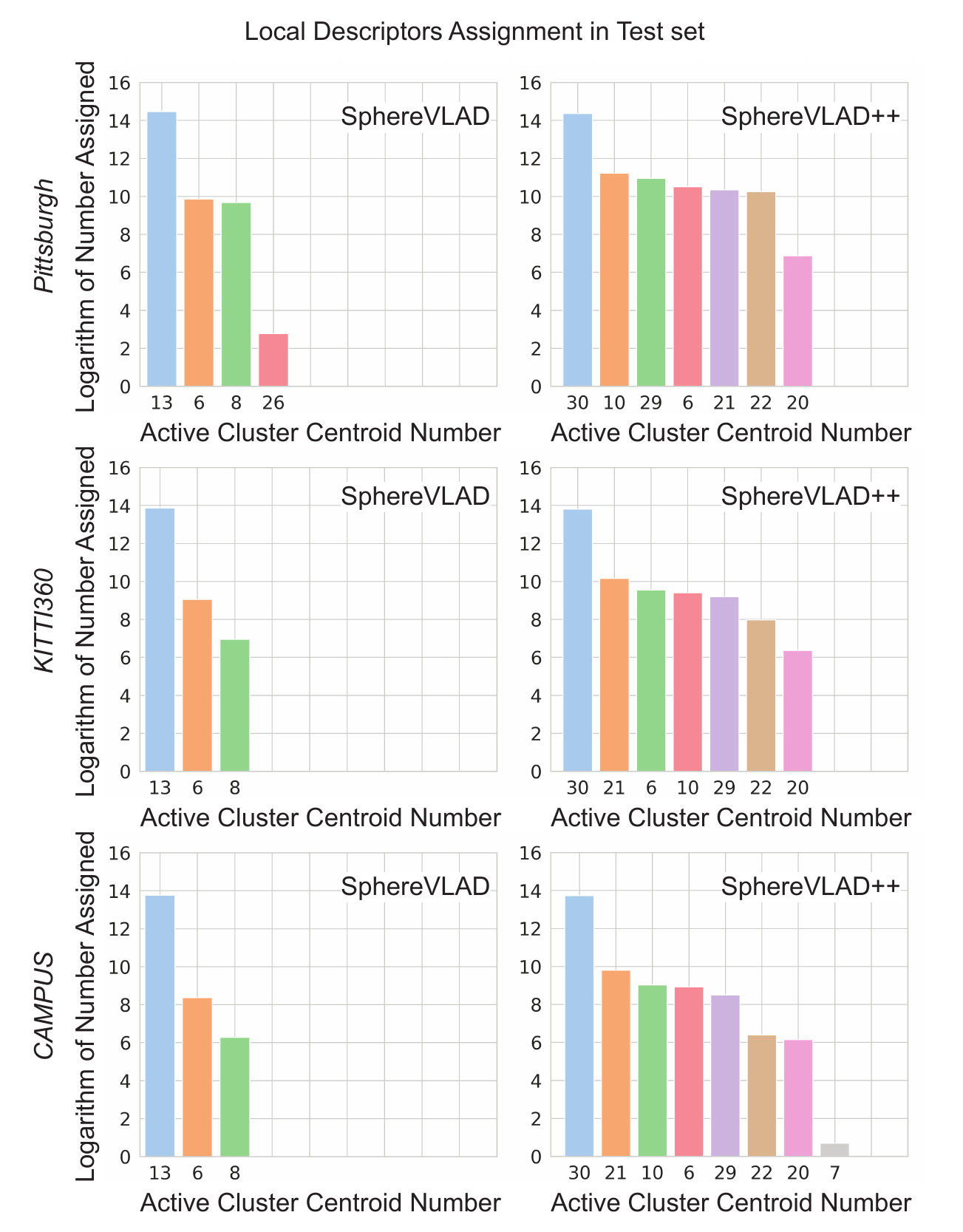}
        \caption{Cluster Assignment Summary on Three Datasets. We select $3824$ samples from 20 trajectories in \textit{Pittsburgh}, $1817$ samples from 8 trajectories in \textit{CMU} and $1967$ sampes from 6 trajectories in \textit{KITTI360} for evaluation.}
        \label{fig:snr_dataset}
    \end{figure}

\subsection{Ablation Study}
    \label{sec:Abalation_study}
    Compared with SphereVLAD, we add two modifications in SphereVLAD++, including the network structure and the parameter settings of the loss function.
    For the network structure, we add a three-dimensional batch normalization after each spherical convolution layer in the feature extractor and an attention module after the feature extractor. 
    To evaluate the contribution of each added structure component, we trained all the models under the same training configuration with SphereVLAD++ and tested them on the \textit{Pittsburgh} dataset.
    The SphereVLAD retrained with the new training parameters is referred to as SphVLAD-new.
    We integrate either the batch normalization or the attention module into SphVLAD-new, denoted as SphVLAD-bn and SphVLAD-Att.
    Finally, we integrate both of two components into SphVLAD-new, denoted as SphVLAD++.
    \begin{table}[ht]
        \centering
        \caption{Contribution of Each Component}
        \begin{tabular}{c c c c}
        \toprule
        Method              & Recall@1          & Recall@1\% \\
        \midrule
        SphVLAD-new         & $82.41$           & $83.34$ \\
        SphVLAD-Att         & $84.72$           & $85.46$ \\
        SphVLAD-bn          & $89.54$           & $90.06$ \\
        SphVLAD++           & $91.11$           & $91.58$ \\
        \bottomrule 
        \end{tabular}
        \label{exp:table_abalation_study}
    \end{table}
    
    As shown in Table.~\ref{exp:table_abalation_study}, the network performance of SphVLAD-Att exceeds the performance of SphVLAD-new by $2.31\%$ and $2.12\%$ on average recall by top $1$ and top $1\%$ respectively, which means our proposed attention module can provide long-range dependencies between local descriptors and therefore benefit global feature aggregation.
    With the batch normalization, SphVLAD-bn sees a significant improvement of $7.13\%$ and $6.34\%$ on average recall, respectively.
    Combining two components can enhance the performance by $8.70\%$ and $8.24\%$, respectively, showing that both components contribute to the improvement.
    
\subsection{Storage and Efficiency Analysis}
    \label{sec:Efficiency_analysis}
    We also analyze the run-time and GPU memory usage between our SphereVLAD++ and other methods.
    The run-time includes the time cost of both preprocessing and inference for each model and we report the averaged results over 500 experiment runs.
    Given the results shown in Table.~\ref{exp:table_runtime}, our method can reach $278hz$ to encode a raw point cloud into a global descriptor.
    Compared with the state-of-art methods, our method takes $3.84ms$ and is only $0.23ms$ slower than the best method SphereVLAD.
    The proprocess of our method containing spherical projection of point cloud and normalization is accelerated by GPU, which only takes $1.6ms$.
    Furthermore, our method is a lightweight model compared with other methods and only requires $1162M$ of GPU usage.
    Therefore, our method is suitable for low-cost mobile robots in long-term navigation.
    \begin{table}[ht]
        \centering
        \caption{Time and Memory Requirements of Methods.}
        \begin{tabular}{c c c c}
        \toprule
        Method &  GPU memory usage  & Run-time per frame \\
        \midrule
        PointNetVLAD &     $1182M$          & $9.57ms$  \\
        LPD-Net      &     $2578M$          & $85.41ms$ \\
        PCAN         &     $7686M$          & $82.07ms$ \\
        SOE-Net      &     $3596M$          & $94.79ms$ \\
        Minkloc3D    &     $1246M$          & $15.05ms$ \\
        SphereVLAD   &     $\textbf{1069M}$          & $\textbf{3.61ms}$ \\
        SphereVLAD++(ours) & $1162M$  & $3.84ms$ \\
        \bottomrule 
        \end{tabular}
        \label{exp:table_runtime}
    \end{table}

    \newpage

\section{Conclusions}
\label{sec:conclusions}
    In this paper, we proposed an attention enhanced, low-cost and viewpoint invariant 3D place recognition method. 
    We develop an attention module which takes advantages of contextual information in order to enhance local features.
    The results on both the public and self-recorded datasets show superiority of our method compared with state-of-the-art in 3D point-cloud retrieval task, especially under large rotation differences, and the contribution of attention module to the global descriptor generator is also illustrated.
    Moreover, the lightweight network structure of our method also enables feasibility of large-scale re-localization task on low-cost mobile robots.
    One notable limitation roots in spherical convolution: the computation cost leads to the low resolution input panorama and shallow network model.
    Moreover, the current version of SNR metric can only evaluate networks using the NetVLAD layer and cannot provide a general metrics for all models.
    In future work, we will improve the retrieval ability by alleviating this limitation and make SNR metric applicable to every model.

    \bibliographystyle{IEEEtran}
    \bibliography{bible}
\endgroup
\end{document}